\documentclass[prodmode,acmtist]{acmsmall} 
\usepackage{bm}
\usepackage{amsmath}
\usepackage{amssymb}
\usepackage[ruled]{algorithm2e}
\usepackage{multirow}

\SetAlFnt{\small}
\SetAlCapFnt{\small}
\SetAlCapNameFnt{\small}
\SetAlCapHSkip{0pt}
\IncMargin{-\parindent}

\begin{document}

\markboth{Y. Wei et al.}{Modality-dependent Cross-media Retrieval}

\title{Modality-dependent Cross-media Retrieval}
\author{YUNCHAO WEI
\affil{Beijing Jiaotong University}
YAO ZHAO
\affil{Beijing Jiaotong University}
ZHENFENG ZHU
\affil{Beijing Jiaotong University}
SHIKUI WEI
\affil{Beijing Jiaotong University}
YANHUI XIAO
\affil{Beijing Jiaotong University}
JIASHI FENG
\affil{University of California, Berkeley}
SHUICHENG YAN
\affil{National University of Singapore}}

\begin{abstract}
In this paper, we investigate the cross-media retrieval between images and text, i.e., using image to search text (I2T) and using text to search images (T2I). Existing cross-media retrieval methods usually learn one couple of projections, by which the original features of images and text can be projected into a common latent space to measure the content similarity. However, using the same projections for the two different retrieval tasks (I2T and T2I) may lead to a tradeoff between their respective performances, rather than their best performances. Different from previous works, we propose a modality-dependent cross-media retrieval (MDCR) model, where two couples of projections are learned for different cross-media retrieval tasks instead of one couple of projections. Specifically, by jointly optimizing the correlation between images and text and the linear regression from one modal space (image or text) to the semantic space, two couples of mappings are learned to project images and text from their original feature spaces into two common latent subspaces (one for I2T and the other for T2I). Extensive experiments show the superiority of the proposed MDCR compared with other methods. In particular, based the 4,096 dimensional convolutional neural network (CNN) visual feature and 100 dimensional LDA textual feature, the mAP of the proposed method achieves 41.5\%, which is a new state-of-the-art performance on the Wikipedia dataset.
\end{abstract}




\acmformat{
}

\begin{bottomstuff}
This work is supported in part by National Basic Research Program of China (No.2012CB316400) and Fundamental Scientific Research Project (No.K15JB00360).

Authors' addresses: Yunchao~Wei, Yao~Zhao, Zhenfeng~Zhu, Shikui~Wei and Yanhui~Xiao are with the Institute of Information Science, Beijing Jiaotong University, Beijing 100044, China, and with the Beijing Key Laboratory of Advanced Information Science and Network Technology, Beijing 100044, China; email:wychao1987@gmail.com, \{yzhao, zhfzhu, shkwei\}@bjtu.edu.cn, xiaoyanhui@gmail.com. Shikui~Wei is also with Hubei Key Laboratory of Intelligent Vision Based Monitoring for Hydroelectric Engineering, China Three Gorges University, Yichang, Hubei 443002, China. Jiashi~Feng is with Department of Electrical Engineering and Computer Sciences, University of California, Berkeley; email:jshfeng@gmail.com. Shuicheng~Yan is with Department of Electrical and Computer Engineering, National University of Singapore; eleyans@nus.edu.sg. 
\end{bottomstuff}

\maketitle

\section{Introduction}

\begin{figure}[t]
	\centering
	\includegraphics[scale=0.7]{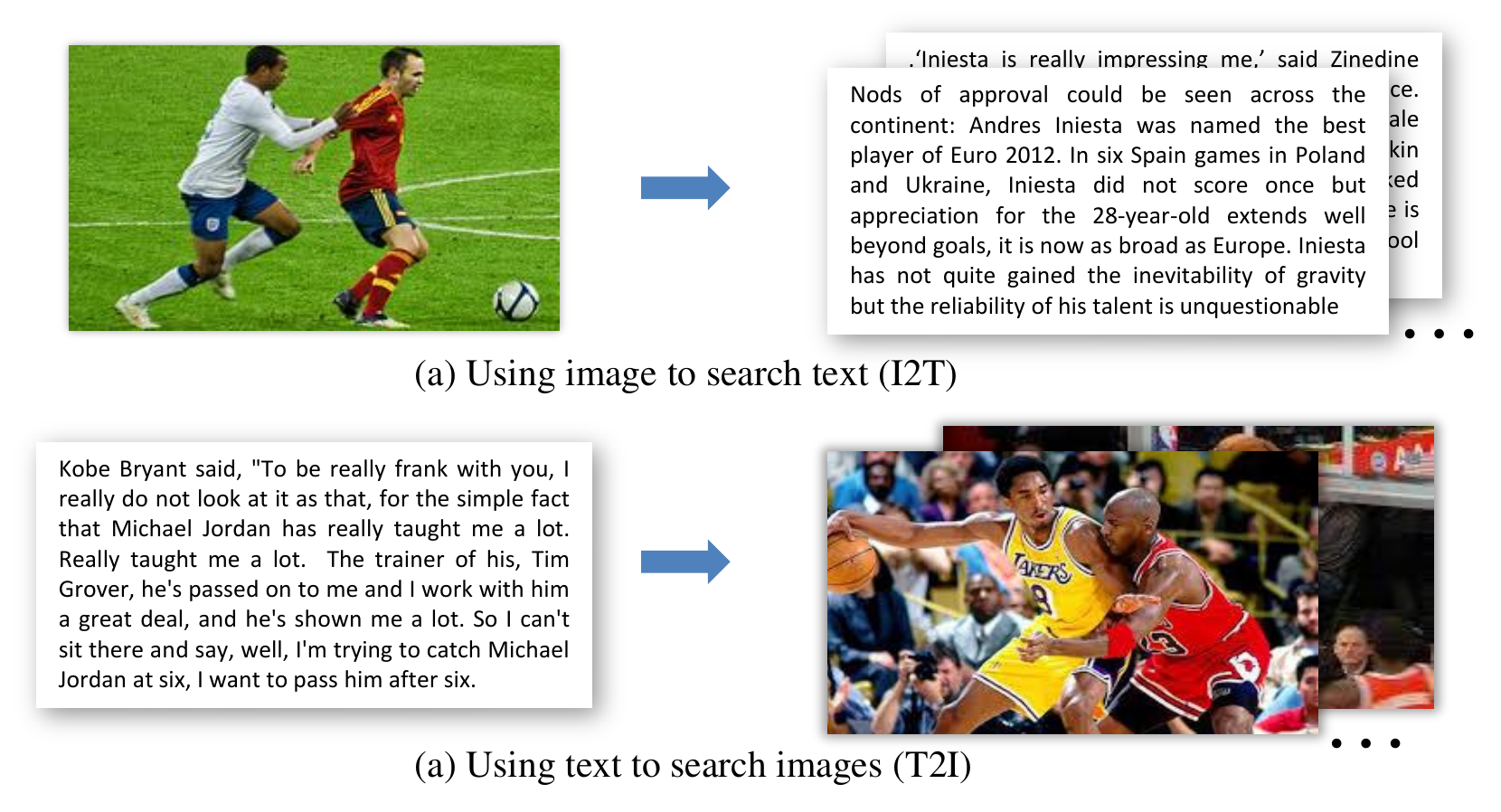}
	\caption{Cross-media retrieval tasks considered in this paper. (a) Given an image of \emph{Iniesta}, the task is to find some text reports related to this image. (b) Given a text document about \emph{Kobe Bryant} and \emph{Michael Jordan}, the task is to find some related images about them. Source images, \copyright ferhat\_culfaz: https://goo.gl/of54g4, \copyright Basket Streaming: https://goo.gl/DfZLRs, \copyright Wikipedia: http://goo.gl/D6RYkt.}
	\label{fig:cro-tasks}
\end{figure}

With the rapid development of information technology, multi-modal data (e.g., image, text, video or audio) have been widely available on the Internet. For example, an image often co-occurs with text on a web page to describe the same object or event. Related research has been conducted incrementally in recent decades, among which the retrieval across different modalities has attracted much attention and benefited many practical applications. However, multi-modal data usually span different feature spaces. This heterogeneous characteristic poses a great challenge to cross-media retrieval tasks. In this work, we mainly focus on addressing the cross-media retrieval between text and images (Fig.~\ref{fig:cro-tasks}), i.e., using image (text) to search text documents (images) with the similar semantics.

To address this issue, many approaches have been proposed by learning a common representation for the data of different modalities. We observe that most exiting works~\cite{2004-Hardoon,2010-NR,2012-Sharma,2013-Gong} focus on learning one couple of mapping matrices to project high-dimensional features from different modalities into a common latent space. By doing this, the correlations of two variables from different modalities can be maximized in the learned common latent subspace. However, only considering pair-wise closeness~\cite{2004-Hardoon} is not sufficient for cross-media retrieval tasks, since it is required that multi-modal data from the same semantics should be united in the common latent subspace. Although~\cite{2012-Sharma} and~\cite{2013-Gong} have proposed to use supervised information to cluster the multi-modal data with the same semantics, learning one couple of projections may only lead to compromised results for each retrieval task. 

\begin{figure*}[t]
	\centering
	\includegraphics[scale=0.6]{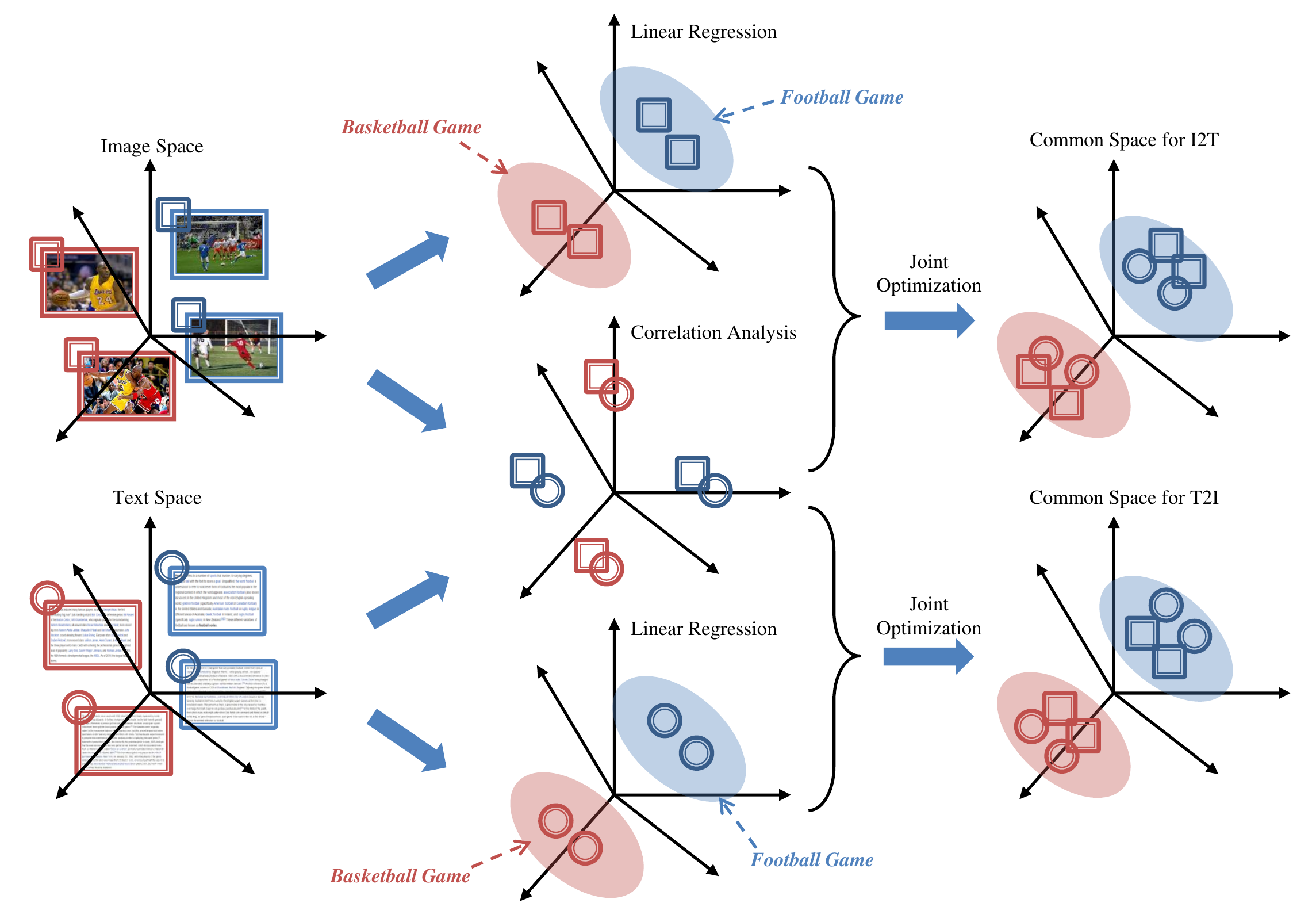}
	\caption{Modality-dependent cross-media retrieval (MDCR) model proposed in this paper. Images are represented by square icons, while text is represented by round icons; different colors indicate different classes. Ellipse fields with blue color and red color indicate semantic clusters of $Football Game$ and $Basketball Game$, respectively. (a) Linear regression from image feature space to semantic space to produce a better separation for images of different classes. (b) Correlation analysis between images and text to keep pair-wise closeness. (c) Linear regression from text feature space to semantic space to produce a better separation for text of different classes. Source images, \copyright Basket Streaming: https://goo.gl/DfZLRs, \copyright Wikipedia: http://goo.gl/RqWL6O, \copyright Wikipedia: http://goo.gl/k3cPs8, \copyright Wikipedia: https://goo.gl/RdgsNL.}
	\label{fig:framwork}
\end{figure*}

In this paper, we propose a modality-dependent cross-media retrieval (MDCR) method, which recommends different treatments for different retrieval tasks, i.e., I2T and T2I. Specifically, MDCR is a task-specific method, which learns two couples of projections for different retrieval tasks. The proposed method is illustrated in Fig.~\ref{fig:framwork}. Fig.~\ref{fig:framwork}(a) and Fig.~\ref{fig:framwork}(c) are two linear regression operations from the image and the text feature space to the semantic space, respectively. By doing this, multi-modal data with the same semantics can be united in the common latent subspace. Fig.~\ref{fig:framwork}(b) is a correlation analysis operation to keep pair-wise closeness of multi-modal data in the common space. We combine Fig.~\ref{fig:framwork}(a) and Fig.~\ref{fig:framwork}(b) to learn a couple of projections for I2T, and a different couple of projections for T2I is jointly optimized by Fig.~\ref{fig:framwork}(b) and Fig.~\ref{fig:framwork}(c). The reason why we learn two couples of projections rather than one couple for different retrieval tasks can be explained as follows. For I2T, we argue that the accurate representation of the query (i.e., the image) in the semantic space is more important than that of the text to be retrieved. If the semantics of the query is misjudged, it will be even harder to retrieve the relevant text. Therefore, only the linear regression term from image feature to semantic label vector and the correlation analysis term are considered for optimizing the mapping matrices for I2T. For T2T, the reason is the same as that for I2T. The main contributions of this work are listed as follow:
\begin{itemize}
\item[$\bullet$] We propose a modality-dependent cross-media retrieval
method, which projects data of different modalities into a common space so that similarity measurement such as Euclidean distance could be applied for cross-media retrieval.
\item[$\bullet$] To better validate the effectiveness of our proposed MDCR, we compare it with other state-of-the-arts based on more powerful feature representations. In particular, with the 4,096 dimensional CNN visual feature and 100 dimensional LDA textual feature, the mAP of the proposed method reaches 41.5\%, which is a new state-of-the-art performance on the Wikipedia dataset as far as we know.
\item[$\bullet$] Based on the INRIA-Websearch dataset~\cite{2010-KAVJ10}, we construct a new dataset for cross-media retrieval evaluation. In addition, all the features utilized in this paper are publicly available\footnote{https://sites.google.com/site/yunchaosite/mdcr}.
\end{itemize}

The remainder of this paper in organized as follows. We briefly review the related work of cross-media retrieval in Section~\ref{sec:rw}. In Section \ref{sec:MDCR}, the proposed modality-dependent cross-media retrieval method is described in detail. Then in Section \ref{sec:experiment}, experimental results are reported and analyzed. Finally, Section \ref{sec:con} presents the conclusions.
\section{Related work}
\label{sec:rw}
During the past few years, numerous methods have been proposed to address cross-media retrieval. 
Some works~\cite{2004-Hardoon,2000-Tenenbaum,2006-Rosipal,2008-yang,2011-Sharma,2010-Hwang,2010-NR,2012-Sharma,2013-Gong,wei2014learning,zhang2014mining} try to learn an optimal common latent subspace for multi-modal data. This kind of methods projects representations of multiple modalities into an isomorphic space, such that similarity measurement can be directly applied between multi-modal data. Two popular approaches, Canonical Correlation Analysis (CCA)~\cite{2004-Hardoon} and Partial Least Squares (PLS)~\cite{2006-Rosipal,2011-Sharma}, are usually employed to find a couple of mappings to maximize the correlations between two variables. Based on CCA, a number of successful algorithms have been developed for cross-media retrieval tasks~\cite{2010-Rashtchian,2010-Hwang,2012-Sharma,2013-Gong}. The work~\cite{2010-Rashtchian} investigated the cross-media retrieval problem in terms of correlation hypothesis and abstraction hypothesis. Based on the isomorphic feature space obtained from CCA, a multi-class logistic regression is applied to generate a common semantic space for cross-media retrieval tasks. In~\cite{2010-Hwang}, Hwang \emph{et al.} used KCCA to develop a cross-media retrieval method by modeling the correlation between visual features and textual features. The work~\cite{2012-Sharma} presented a generic framework for multi-modal feature extraction techniques, called Generalized Multiview Analysis~(GMA). More recently, the work~\cite{2013-Gong} proposed a three-view CCA model by introducing a semantic view to produce a better separation for multi-modal data of different classes in the learned latent subspace.

To address the problem of prohibitively expensive nearest neighbor search, some hashing-based approaches \cite{2011-kumar,2013-wu-hash} to large scale similarity search have drawn much interest from the cross-media retrieval community. In particular, \cite{2011-kumar} proposed a cross view hashing method to generate hash codes by minimizing the distance of hash codes for the similar data and maximizing the distance for the dissimilar data. Recently, \cite{2013-wu-hash} proposed a sparse multi-modal hashing method, which can obtain sparse codes for the data across different modalities via joint multi-modal dictionary learning, to address cross-modal retrieval. Besides, with the development of deep learning, some deep models~\cite{2013-Devise,2014-wang,2014-lu-learning,2014-zhuang-cross} have also been proposed to address cross-media problems. Specifically, \cite{2013-Devise} presented a deep visual-semantic embedding model to identify visual objects using both labeled image data and semantic information obtained from unannotated text documents. \cite{2014-wang} proposed an effective mapping mechanism, which can capture both intra-modal and inter-modal semantic relationships of multi-modal data from heterogeneous sources, based on the stacked auto-encoders deep model.

Beyond the above mentioned models, some other works~\cite{2009-yang-ranking,2010-yang-query,2012-yang,2013-wu,2013-zhai,2014-cuicui} have also been    proposed to address cross-media problems. In particular, \cite{2013-wu} presented a bi-directional cross-media semantic representation model by  optimizing the bi-directional list-wise ranking loss with a latent space embedding. In \cite{2013-zhai}, both the intra-media and the inter-media correlation are explored for cross-media retrieval. Most recently, \cite{2014-cuicui} presented a heterogeneous similarity learning approach based on metric learning for cross-media retrieval. With the convolutional neural network (CNN) visual feature, some new state-of-the-art cross-media retrieval results have been achieved in \cite{2014-cuicui}.

\section{Modality-dependent Cross-media Retrieval}
\label{sec:MDCR}
In this section, we detail the proposed supervised cross-media retrieval method, which we call modality-dependent cross-media retrieval (MDCR). Each pair of image and text in the training set is accompanied with semantic information (e.g., class labels). Different from~\cite{2013-Gong} which incorporates the semantic information as a third view, in this paper, semantic information is employed to determine a common latent space with a fixed dimension where samples with the same label can be clustered.

Suppose we are given a dataset of $n$ data instances, i.e., $\mathcal {G} = \{({\bf{x}}_i,{\bf{t}}_i)\}_{i=1}^{n}$, where ${\bf{x}}_i\in\mathbb{R}^{p}$ and ${\bf{t}}_i\in\mathbb{R}^{q}$ are original low-level features of image and text document, respectively. Let $X = {[{{\bf{x}}_1},...,{{\bf{x}}_n}]^T} \in {\mathbb{R}^{n \times p}}$ be the feature matrix of image data, and $T = {[{{\bf{t}}_1},...,{{\bf{t}}_n}]^T} \in {\mathbb{R}^{n \times q}}$ be the feature matrix of text data. Assume that there are $c$ classes in $\mathcal {G}$. $S = {[{{\bf{s}}_1},...,{{\bf{s}}_n}]^T} \in {\mathbb{R}^{n \times c}}$ is the semantic matrix with the $i$th row being the semantic vector corresponding to ${\bf{x}}_i$ and ${\bf{t}}_i$. In particular, we set the $j$th element of ${\bf{s}}_i$ as 1, if ${\bf{x}}_i$ and ${\bf{t}}_i$ belong to the $j$th class.

\noindent\textbf{Definition 1:} The cross-media retrieval problem is to learn two optimal mapping matrices $V\in\mathbb{R}^{c\times p}$ and $W\in\mathbb{R}^{c\times q}$ from the multi-modal dataset $\mathcal {G}$, which can be formally formulated into the following optimization framework:
\begin{equation}
\label{eq:fram}
\mathop {\min }\limits_{V,W} \;f(V,W) = \mathcal{C}(V, W) + \mathcal{L}(V, W) + \mathcal{R}(V,W),
\end{equation}
where $f$ is the objective function consisting of three terms. In particular, $\mathcal{C}(V, W)$ is a correlation analysis term used to keep pair-wise closeness of multi-modal data in the common latent subspace. $\mathcal{L}(V, W)$ is a linear regression term from one modal feature space (image or text) to the semantic space, used to centralize the multi-modal data with the same semantics in the common latent subspace. $\mathcal{R}(V, W)$ is the regularization term to control the complexity of the mapping matrices $V$ and $W$.

In the following subsections, we will detail the two algorithms for I2T and T2I based on the optimization framework Eq.(\ref{eq:fram}).

\subsection{Algorithm for I2T}
\label{sec:I2T}
This section addresses the cross-media retrieval problem of using an image to retrieve its related text documents. Denote the two optimal mapping matrices for images and text as $V_1\in\mathbb{R}^{c\times p}$ and $W_1\in\mathbb{R}^{c\times q}$, respectively. Based on the optimization framework Eq.(\ref{eq:fram}), the objective function of I2T is defined as follows:
\begin{equation}
\label{eq:I2T}
\begin{aligned}
\mathop {\min }\limits_{{V_1},{W_1}} \;f\left( {{V_1},{W_1}} \right) = &\lambda \left\| {XV_1^T - TW_1^T} \right\|_F^2 + \left( {1 - \lambda } \right)\left\| {XV_1^T - S} \right\|_F^2\\
&+ R\left( {{V_1},{W_1}} \right),
\end{aligned}
\end{equation}
where $0 \le \lambda  \le 1$ is a tradeoff parameter to balance the importance of the correlation analysis term and the linear regression term, ${\left\| {\; \cdot \;} \right\|_F}$ denotes the Frobenius norm of the matrix, and $R\left( {{V_1},{W_1}} \right)$ is the regularization function used to regularize the mapping matrices. In this paper, the regularization function is defined as:
\[R\left( {{V_1},{W_1}} \right) = {\eta _1}\left\| V_1 \right\|_F^2 + {\eta _2}\left\| W_1 \right\|_F^2,\]
where $\eta_1$ and $\eta_2$ are nonnegative parameters to balance these two regularization terms.

\subsection{Algorithm for T2I}
\label{sec:T2I}
This section addresses the cross-media retrieval problem of using text to retrieve its related images. Different from the objective function of I2T, the linear regression term for T2I is a regression operation from the textual space to the semantic space. Denote the two optimal mapping matrices for images and text in T2I as $V_2\in\mathbb{R}^{c\times p}$ and $W_2\in\mathbb{R}^{c\times q}$, respectively. Based on the optimization framework Eq.(\ref{eq:fram}), the objective function of T2I is defined as follows:
\begin{equation}
\label{eq:T2I}
\begin{aligned}
\mathop {\min }\limits_{{V_2},{W_2}} \;f\left( {{V_2},{W_2}} \right) = &\lambda \left\| {XV_2^T - TW_2^T} \right\|_F^2 + \left( {1 - \lambda } \right)\left\| {TW_2^T - S} \right\|_F^2\\
&+ R\left( {{V_2},{W_2}} \right),
\end{aligned}
\end{equation}
where the setting of the tradeoff parameter $\lambda$ and the regularization function $R\left( {{V_2},{W_2}} \right)$ are consistent with the setting presented in Section~\ref{sec:I2T}.

\subsection{Optimization}
The optimization problems for I2T and T2I are unconstrained optimization with respect to two matrices. Hence, both Eq.(\ref{eq:I2T}) and Eq.(\ref{eq:T2I}) are non-convex optimization problems and only have many local optimal solutions. For the non-convex problem, we usually design algorithms to seek stationary points. We note that Eq.(\ref{eq:I2T}) is convex with respect to either $V_1$ or $W_1$ while fixing the other. Similarly, Eq.(\ref{eq:T2I}) is also convex with respect to either $V_2$ or $W_2$ while fixing the other. Specifically, by fixing $V_1$($V_2$) or $W_1$($W_2$), the minimization over the other can be finished with the gradient descent method.

The partial derivatives of $V_1$ or $W_1$ over Eq.(\ref{eq:I2T}) are given as follows:
\begin{equation}
\label{eq:gradientI2T1}
\begin{array}{l}
{\nabla _{{V_1}}}f\left( {{V_1},{W_1}} \right) = {V_1}{X^T}X + 2\left[ {{\eta _1}{V_1} - \lambda {W_1}{T^T}X - \left( {1 - \lambda } \right){S^T}X} \right],
\end{array}
\end{equation}
\begin{equation}
\label{eq:gradientI2T2}
\begin{array}{l}
{\nabla _{{W_1}}}f\left( {{V_1},{W_1}} \right) = 2\left[ {{\eta _2}{W_1} + \lambda \left( {{W_1}{T^T}T - {V_1}{X^T}T} \right)} \right].
\end{array}
\end{equation}

Similarly, the partial derivatives of $V_2$ or $W_2$ over Eq.(\ref{eq:T2I}) are given as follows:
\begin{equation}
\label{eq:gradientT2I1}
\begin{array}{l}
{\nabla _{{V_2}}}f\left( {{V_2},{W_2}} \right) = 2\left[ {{\eta _1}{V_2} + \lambda \left( {{V_2}{X^T}X - {W_2}{T^T}X} \right)} \right],\\
\end{array} 
\end{equation}
\begin{equation}
\label{eq:gradientT2I2}
\begin{array}{l}
{\nabla _{{W_2}}}f\left( {{V_2},{W_2}} \right) = W{T^T}T + 2\left[ {{\eta _2}{W_2} - \lambda {V_2}{X^T}T - (1 - \lambda ){S^T}T} \right].
\end{array}
\end{equation}

A common way to solve this kind of optimization problems is an alternating updating process until the result converges. Algorithm 1 summarizes the optimization procedure of the proposed MDCR method for I2T, which can be easily extended for T2I. 

\begin{algorithm}[h]
\SetAlgoNoLine
\KwIn{The feature matrix of image data $X = {[{{\bf{x}}_1},...,{{\bf{x}}_n}]^T} \in {\mathbb{R}^{n \times p}}$, the feature matrix of text data $T = {[{{\bf{t}}_1},...,{{\bf{t}}_n}]^T} \in {\mathbb{R}^{n \times q}}$, the semantic matrix corresponding to images and text $S = {[{{\bf{s}}_1},...,{{\bf{s}}_n}]^T} \in {\mathbb{R}^{n \times c}}$.}
Initialize $V_1^{(\upsilon)}$, $W_1^{(\omega)}$, 
$\upsilon\leftarrow$0 and $\omega\leftarrow$0. Set the parameters $\lambda$, $\eta_1$, $\eta_2$, $\mu$ and $\epsilon$. $\mu$ is the step size in the alternating updating process and $\epsilon$ is the convergence condition.\\
\Repeat{Convergence or maximum iteration number achieves.}{
        Alternative optimization process for I2T (Algorithm \ref{alg::I2T}).
        }
     
\caption{Optimization for Modality-dependent Cross-media Retrieval}
\KwOut{$V_1^{(\upsilon)}$, $W_1^{(\omega)}$. 
}
\label{alg::fram}
\end{algorithm}

\begin{algorithm}[h]
\Repeat{$value1 - value2\leq\epsilon$}{
	Set $value1=f\left( {V_1^{(\upsilon )},W_1^{(\omega )}} \right)$\;
	Update $V_1^{(\upsilon  + 1)} = V_1^{(\upsilon )} - \mu {\nabla _{V_1^{(\upsilon )}}}f\left( {V_1^{(\upsilon )},W_1^{(\omega )}} \right)$\;
	Set $value2=f\left( {V_1^{(\upsilon + 1 )},W_1^{(\omega )}} \right)$, $\upsilon\leftarrow\upsilon + 1$\;}
\Repeat{$value1 - value2\leq\epsilon$}{
	Set $value1=f\left( {V_1^{(\upsilon )},W_1^{(\omega )}} \right)$\;
	Update $W_1^{(\omega  + 1)} = W_1^{(\omega )} - \mu {\nabla _{W_1^{(\omega )}}}f\left( {V_1^{(\upsilon )},W_1^{(\omega )}} \right)$\;
	Set $value2=f\left( {V_1^{(\upsilon )},W_1^{(\omega + 1 )}} \right)$, $\omega\leftarrow\omega + 1$\;}
\caption{Alternative Optimization Process for I2T}
\label{alg::I2T}
\end{algorithm}

\section{Experimental Results}
\label{sec:experiment}
To evaluate the proposed MDCR algorithm, we systematically compare it with other state-of-the-art methods on three datasets, i.e., Wikipedia~\cite{2010-NR}, Pascal Sentence~\cite{2010-Rashtchian} and a subset of INRIA-Websearch~\cite{2010-KAVJ10}.

\subsection{Datasets}
\noindent\textbf{Wikipedia\footnote{http://www.svcl.ucsd.edu/projects/crossmodal/}:} This dataset contains totally 2,866 image-text pairs from 10 categories. The whole dataset is randomly split into a training set and a test set with 2,173 and 693 pairs. We utilize the publicly available features provided by~\cite{2010-NR} i.e., 128 dimensional SIFT BoVW for images and 10 dimensional LDA for text, to compare directly with existing results. Besides, we also present the cross-media retrieval results based on the 4,096 dimensional CNN visual features\footnote{The CNN model is pre-trained on ImageNet. We utilize the outputs from the second fully-connected layer as the CNN visual feature in this paper. For more details, please refer to~\cite{2012-krizhevsky}.}  and the 100 dimensional Latent Dirichlet Allocation model (LDA)~\cite{2003-Blei} textual features (we firstly obtain the textual feature vector based on 500 tokens and then LDA model is used to compute the probability of each document under 100 topics).

\noindent\textbf{Pascal Sentence\footnote{http://vision.cs.uiuc.edu/pascal-sentences/}:} This dataset contains 1,000 pairs of image and text descriptions from 20 categories (50 for each category). We randomly select 30 pairs from each category as the training set and the rest are taken as the testing set. We utilize the 4,096 dimensional CNN visual feature for image representation.  For textual features, we firstly extract the feature vector based on 300 most frequent tokens (with stop words removed) and then utilize the LDA to compute the probability of each document under 100 topics. The 100 dimensional probability vector is used for textual representation.

\noindent\textbf{INRIA-Websearch:} This dataset contains 71,478 pairs of image and text annotations from 353 categories. We remove those pairs which are marked as \emph{irrelevant}, and select those pairs that belong to any one of the 100 largest categories. Then, we get a subset of 14,698 pairs for evaluation. We randomly select 70\% pairs from each category as the training set (10,332 pairs), and the rest are treated as the testing set (4,366 pairs). We utilize the 4,096 dimensional CNN visual feature for image representation. For textual features, we firstly obtain the feature vector based on 25,000 most frequent tokens (with stop words removed) and then employ the LDA to compute the probability of each document under 1,000 topics. 

For semantic representation, the ground-truth labels of each dataset are employed to construct semantic vectors (10 dimensions for Wikipedia dataset, 20 dimensions for Pascal Sentence dataset, and 100 dimensions for INRIA-Websearch dataset) for pairs of image and text. 

\subsection{Experimental Settings}
In the experiment, Euclidean distance is used to measure the similarity between features in the embedding latent subspace. Retrieval performance is evaluated by mean average precision (mAP), which is one of the standard information retrieval metrics. Specifically, given a set of queries, the average precision (AP) of each query is defined as:
\[AP = \frac{{\sum\nolimits_{k = 1}^R {P(k)rel(k)} }}{{\sum\nolimits_{k = 1}^R {rel(k)} }},\]
where $R$ is the size of the test dataset. $rel(k)=1$ if the item at rank $k$ is relevant, $rel(k)=0$ otherwise. $P(k)$ denotes the precision of the result ranked at $k$. We can get the mAP score by averaging AP for all queries. 

\begin{table}[tbp]
	\begin{center}
		\tbl{mAP scores for image and text query on the Wikipedia dataset based on the publicy available featrues.\label{tab:wiki}}{
			
			\begin{tabular}{|c|c|c|c|c|c|c|c|c|c|}
				\hline
				Query &        PLS &        BLM &        CCA &         SM &        SCM &      GMMFA &      GMLDA &   T-V CCA   &MDCR \\
				\hline
				Image &      0.207 &      0.237 &      0.182 &      0.225 &      0.277 &      0.264 &      0.272 &      0.228 & {\bf 0.287} \\
				\hline
				Text &      0.192 &      0.144 &      0.209 &      0.223 &      0.226 &      0.231 & {\bf 0.232} &       0.205 &     0.225 \\
				\hline
				Average &      0.199 &      0.191 &      0.196 &      0.224 &      0.252 &      0.248 &      0.253 &      0.217 & {\bf 0.256} \\
				\hline
			\end{tabular}}
		\end{center}
	\end{table}
\subsection{Results}
In the experiments, we mainly compare the proposed MDCR with six algorithms, including CCA, Semantic Matching (SM)~\cite{2010-NR}, Semantic Correlation Matching (SCM)~\cite{2010-NR}, Three-View CCA (T-V CCA)~\cite{2013-Gong}, Generalized Multiview Marginal Fisher Analysis (GMMFA)~\cite{2012-Sharma} and Generalized Multiview Linear Discriminant Analysis (GMLDA)~\cite{2012-Sharma}.

For the Wikipedia dataset, we firstly compare the proposed MDCR with other methods based on the publicly available features~\cite{2010-NR}, i.e., 128-SIFT BoVW for images and 10-LDA for text. We fix $\mu$ = $0.02$ and $\epsilon$ = ${10^{ - 4}}$, and experimentally set $\lambda = 0.1$, $\eta_1 = 0.5$ and $\eta_2 = 0.5$ for the optimization of I2T, and the parameters for T2I are set as $\lambda = 0.5$, $\eta_1 = 0.5$ and $\eta_2 = 0.5$. The mAP scores for each method are shown in Table~\ref{tab:wiki}. It can be seen that our method is more effective compared with other common space learning methods. To further validate the necessity to be task-specific for cross-media retrieval, we  evaluate the proposed method in terms of training a unified $V$ and $W$ by incorporating both two linear regression terms in Eq.(\ref{eq:I2T}) and Eq.(\ref{eq:T2I}) into a single optimization objective. As shown in Table~\ref{tab:wiki2}, the learned subspaces for I2T and T2I could not be used interchangeably and the unified scheme can only achieve compromised performance for each retrieval task, which cannot compare to the proposed modality-dependent scheme.

\begin{table}[htbp]
	\begin{center}
		\tbl{Comparison between MDCR and its unified scheme for cross-media retrieval on the Wikipedia dataset.\label{tab:wiki2}}{
			\begin{tabular}{|c|c|c|c|}
				\hline
				Wikipedia & MDCR-Eq.(\ref{eq:I2T}) & MDCR-Eq.(\ref{eq:T2I}) & Unified Scheme \\
				\hline
				I2T   & {\bf 0.287} & 0.165 & 0.236 \\
				\hline
				T2I   & 0.146 & {\bf 0.225} & 0.216 \\
				\hline
			\end{tabular}}%
		\end{center}
	\end{table}%

\begin{figure}
	\centering
	\includegraphics[scale=0.55]{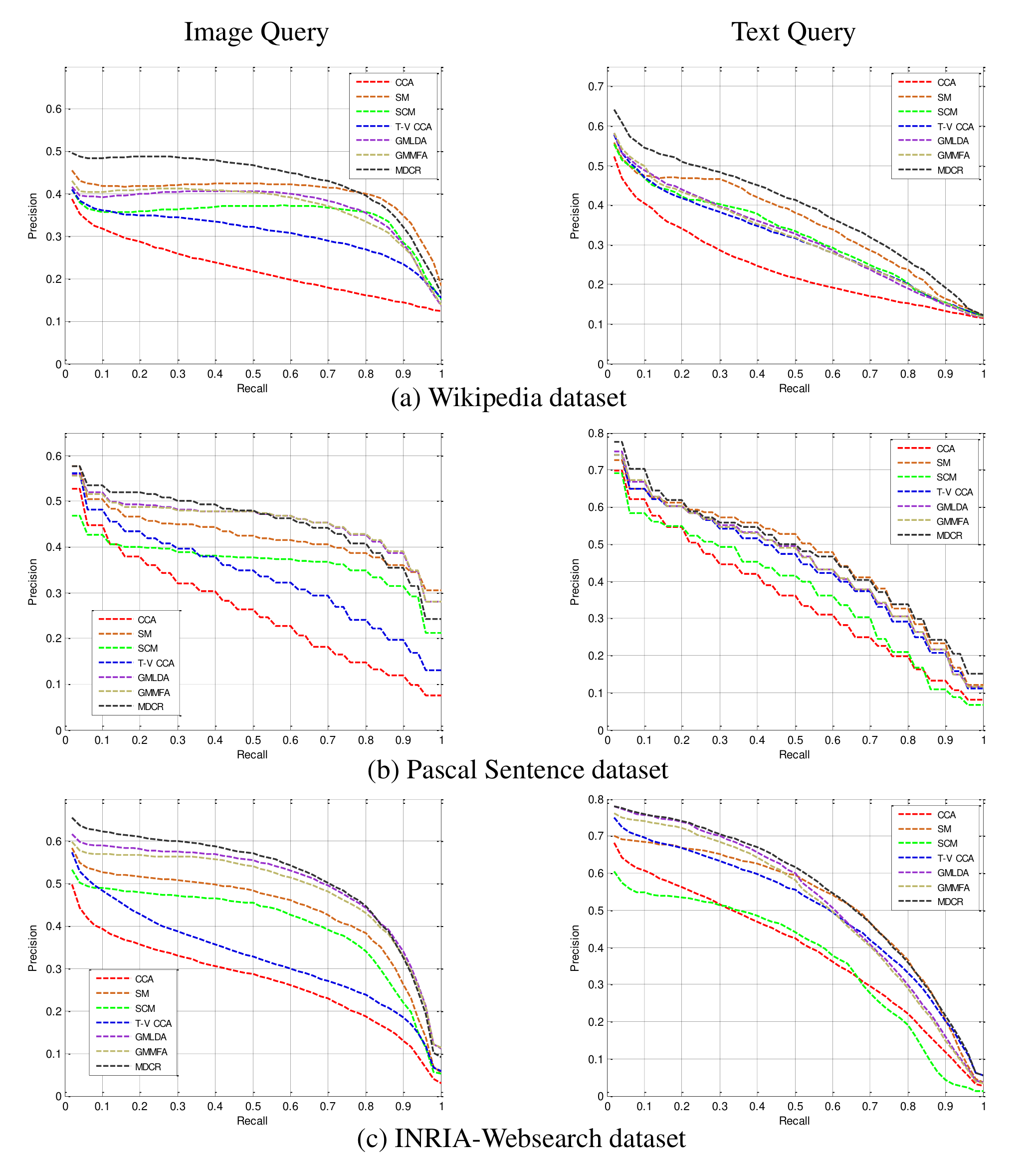}
	\caption{Precision-Recall curves of the proposed MDCR and compared methods}
	\label{fig:cnn_roc_com}
\end{figure}

\begin{figure}[t]
	\centering
	\includegraphics[scale=0.55]{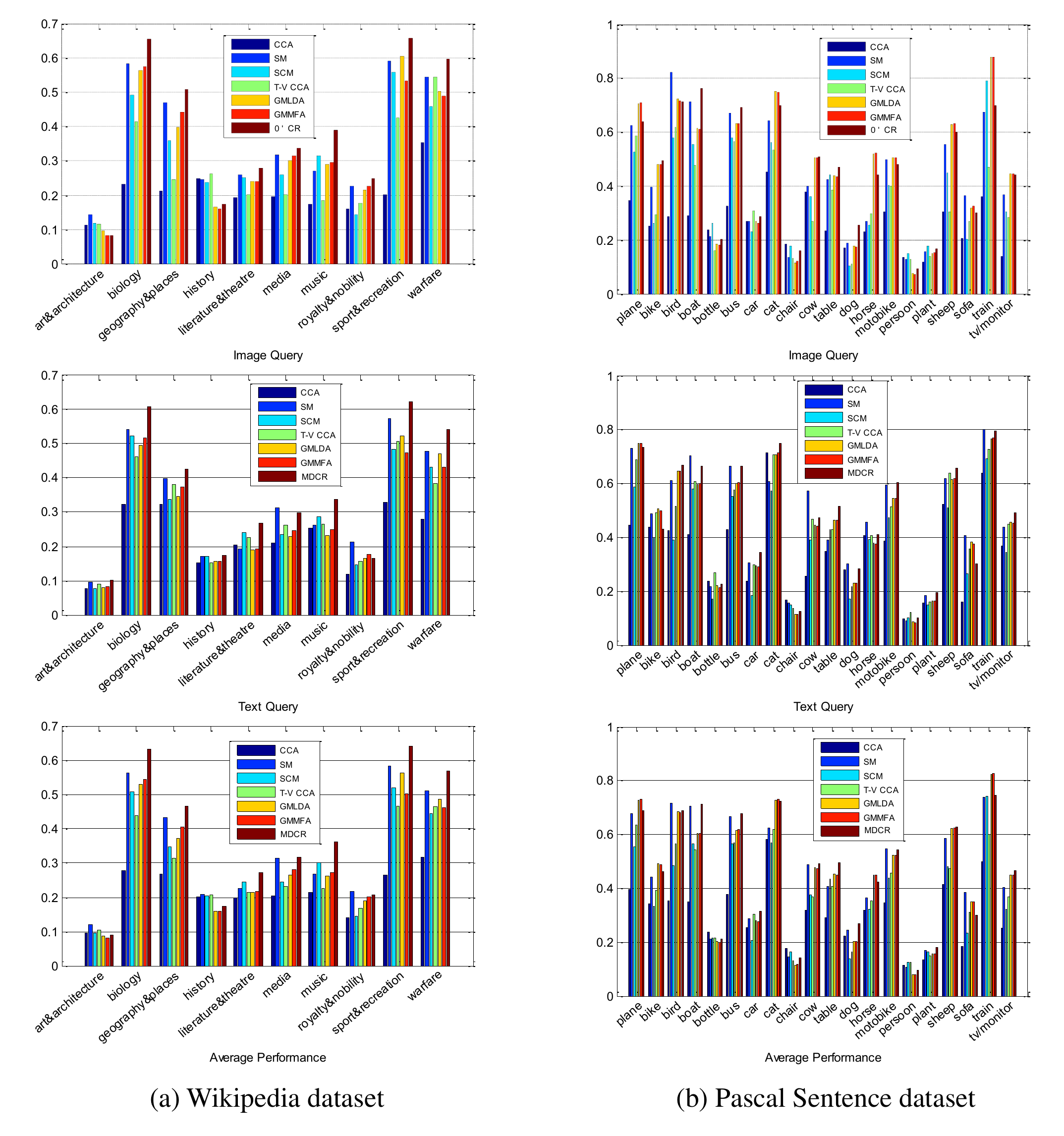}
	\caption{mAP performance for each class on the Wikipedia dataset and the Pascal Sentence dataset.}
	\label{fig:cnn_bar_com}
\end{figure}

\begin{figure}
	\centering
	\includegraphics[scale=0.55]{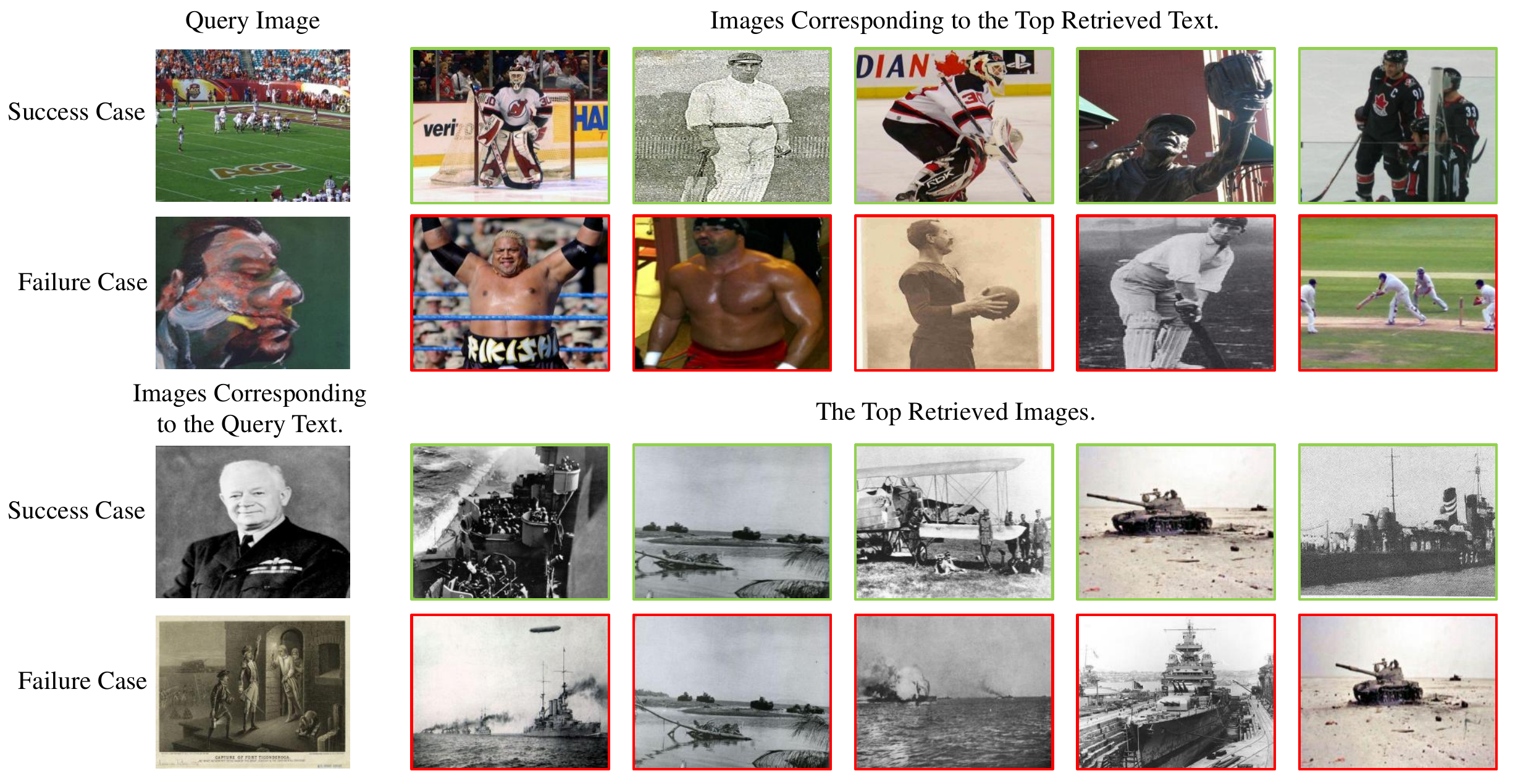}
	\caption{Some successful and failure cases of our method on the Wikipedia dataset. Green and red borders indicate true and false retrieval results, respectively. All the images in this figure are from the Wikipedia dataset~\cite{2010-NR}.}
	\label{fig:example}
\end{figure}

As a very popular dataset,  Wikipedia has been employed by many other works for cross-media retrieval evaluation. With a different \emph{train/test} division, \cite{2013-wu-hash} achieved an average mAP score of 0.226 (Image Query: 0.227, Text Query: 0.224) through a sparse hash model and \cite{2014-wang} achieved an average mAP score of 0.183 (Image Query: 0.187, Text Query: 0.179) through a deep auto-encoder model. Besides, some other works utilized their own extracted features (both for images and text) for cross-media retrieval evaluation. To further validate the effectiveness of the proposed method, we also compare MDCR with other methods based on more powerful features, i.e., 4,096-CNN for images and 100-LDA for text. We fix $\mu$ = $0.02$ and $\epsilon$ = ${10^{ - 4}}$, and experimentally set $\lambda = 0.1$, $\eta_1 = 0.5$ and $\eta_2 = 0.5$ for the optimization of I2T and T2I. The comparison results are shown in Table~\ref{tab:cnn_com}. It can be seen that some new state-of-the-art performances are achieved by these methods based on the new feature representations and the proposed MDCR can also outperform others. In addition, we also compare our method with the recent work~\cite{2014-cuicui}, which utilizes 4,096-CNN for images and 200-LDA for text, in Table~\ref{tab:wikicnn_com}. We can see that the proposed MDCR reaches a new state-of-the-art performance on the Wikipedia dataset. Please refer to Fig.~\ref{fig:cnn_roc_com} for the comparisons of Precision-Recall curves and Fig.~\ref{fig:cnn_bar_com} for the mAP score of each category. Figure~\ref{fig:example} gives some successful and failure cases of our method. For the image query (the 2nd row), although the query image is categorized into \emph{Art}, it is prevailingly characterized by the human figure, i.e., a strong man, which has been captured by our method and thus leads to the failure results shown. For the text query (the 4th row), there exist many \emph{Warfare} descriptions in the document such as \emph{war, army} and \emph{troops}, which can be hardly realted to the label of the query text, i.e.  \emph{Art}. 
\begin{table}	
	\begin{center}
		\tbl{Cross-media retrieval comparition with results of four methods reported by~\cite{2014-cuicui} on the Wikipedia dataset.\label{tab:wikicnn_com}}{
			\begin{tabular}{|c|c|c|c|c|c|}
				\hline
				Query &        \emph{GMLDA} &       \emph{GMMFA} &        \emph{MsAlg} &         \emph{LRBS} &     TSCR  \\ \hline
				Image &        0.368 &	      0.387	&        0.373 &   \bf{0.445} &     0.435 \\ \hline		 
				Text  &        0.297 &	      0.311	&        0.327 &	    0.377 &	\bf{0.394} \\ \hline		 
				Average &      0.332 &	      0.349	&        0.350 &	    0.411 &	 \bf{0.415} \\ \hline	    	 
			\end{tabular}}
		\end{center}
	\end{table}

For the Pascal Sentence dataset and the INRIA-Websearch dataset, we experimentally set $\lambda = 0.5$, $\eta_1 = 0.5$, $\eta_2 = 0.5$, $\mu$ = $0.02$ and $\epsilon$ = ${10^{ - 4}}$ during the alternative optimization process for I2T and T2T. The comparison results can be found in Table~\ref{tab:cnn_com}. It can be seen that our method is more effective compared with others even on a more challenging dataset, i.e., INRIA-Websearch (with 14,698 pairs of multi-media data and 100 categories). Please refer to Fig.~\ref{fig:cnn_roc_com} for the comparisons of Precision-Recall curves for these two datasets and Fig.~\ref{fig:cnn_bar_com} for the mAP score of each category on the Pascal Sentence dataset.

\begin{table}[h]
\begin{center}
\tbl{Comparitions of cross-media retrieval performance.\label{tab:cnn_com}}{
	
	\begin{tabular}{|c|c|c|c|c|c|c|c|c|}
		\hline
		Dataset                 & Query   & CCA   & SM    & SCM   & T-V CCA & GMLDA & GMMFA & MDCR  \\ \hline\hline
		\multirow{3}{*}{Wikipedia}   & Image   & 0.226 & 0.403 & 0.351 & 0.310 & 0.372 & 0.371  & \bf{0.435} \\ \cline{2-9} 
		& Text    & 0.246 & 0.357 & 0.324 & 0.316 & 0.322 & 0.322  & \bf{0.394} \\ \cline{2-9} 
		& Average & 0.236 & 0.380 & 0.337 & 0.313 & 0.347 & 0.346  & \bf{0.415} \\ \hline \hline
		\multirow{3}{*}{Pascal Sentence} & Image   & 0.261 & 0.426 & 0.369 & 0.337 & \bf{0.456} & 0.455  & 0.455 \\ \cline{2-9} 
		& Text    & 0.356 & 0.467 & 0.375 & 0.439 & 0.448 & 0.447  & \bf{0.471} \\ \cline{2-9} 
		& Average & 0.309 & 0.446 & 0.372 & 0.388 & 0.452 & 0.451  & \bf{0.463} \\ \hline \hline
		\multirow{3}{*}{INRIA-Websearch}  & Image   & 0.274 & 0.439 & 0.403 & 0.329 & 0.505 & 0.492  & \bf{0.520} \\ \cline{2-9} 
		& Text    & 0.392 & 0.517 & 0.372 & 0.500 & 0.522 & 0.510  & \bf{0.551} \\ \cline{2-9} 
		& Average & 0.333 & 0.478 & 0.387 & 0.415 & 0.514 & 0.501 & \bf{0.535} \\ \hline
	\end{tabular}}
\end{center}
\end{table}

\section{Conclusions}
\label{sec:con}
Cross-media retrieval has long been a challenge. In this paper, we focus on designing an effective cross-media retrieval model for images and text, i.e., using image to search text (I2T) and using text to search images (T2I). Different from traditional common space learning algorithms, we propose a modality-dependent scheme which recommends different treatments for I2T and T2I by learning two couples of projections for different cross-media retrieval tasks. Specifically, by jointly optimizing a correlation term (between images and text) and a linear regression term (from one modal space, i.e., image or text to the semantic space), two couples of mappings are gained for different retrieval tasks. Extensive experiments on the Wikipedia dataset, the Pascal Sentence dataset and the INRIA-Websearch dataset show the superiority of the proposed method compared with state-of-the-arts.

\bibliographystyle{ACM-Reference-Format-Journals}
\bibliography{acmsmall-sample-bibfile}

\received{April 2014}{January 2015}{May 2015}
%


\end{document}